\documentclass[sigconf, nonacm, screen]{acmart}

\newcommand\vldbdoi{XX.XX/XXX.XX}

\newcommand\vldbavailabilityurl{}
\newcommand\vldbpagestyle{plain} 

\usepackage{booktabs} 
\usepackage{amsmath}
\usepackage{graphicx}
\usepackage{wrapfig}
\usepackage{subfigure}
\usepackage{subcaption}
\usepackage[np,autolanguage]{numprint}
\usepackage{units}
\usepackage{fontawesome}
\usepackage{balance}
\usepackage{algorithm}
\usepackage[noend]{algpseudocode}
\usepackage{listings}
\usepackage{color}
\usepackage{appendix}
\usepackage{enumitem}
\usepackage{makecell}
\usepackage{cleveref}

\definecolor{dkgreen}{rgb}{0,0.6,0}
\definecolor{gray}{rgb}{0.5,0.5,0.5}
\definecolor{mauve}{rgb}{0.58,0,0.82}
\definecolor{dkred}{rgb}{1.0,0.01,0.24}

\newcommand{\header}[1]{\vspace{0.7mm}\noindent\textbf{#1}.}
\newcommand{\headerl}[1]{\vspace{1mm}\noindent\textit{#1}.}
\newcommand{\headerul}[1]{\vspace{1mm}\noindent\textit{\underline{#1}}.}

\usepackage{tikz}
\newcommand*\circled[1]{%
\tikz[baseline=(char.base)]{%
\node[shape=circle,draw,fill=black,text=white,inner sep=1.2pt] (char) {#1};}}

\setlist[itemize]{topsep=2pt,itemsep=0pt,parsep=0pt,partopsep=0pt} 

\begin{document}
\title{Be Fair! Can Machine Learning Engineering Agents Adhere~to~Fairness~Constraints?}

\author{Anna Richter}
\affiliation{\institution{BIFOLD \& TU Berlin}}
\email{a.richter@tu-berlin.de}

\author{Julia Stoyanovich}
\affiliation{\institution{New York University}}
\email{stoyanovich@nyu.edu}

\author{Sebastian Schelter}
\affiliation{\institution{BIFOLD \& TU Berlin}}
\email{schelter@tu-berlin.de}

\renewcommand{\shortauthors}{Anna Richter et al.}

\begin{abstract}
Machine learning engineering (MLE) agents promise to automate end-to-end ML pipeline development from raw data and natural language instructions, potentially making ML accessible to non-technical domain experts. However, in sensitive and regulated domains, this abstraction creates a responsibility gap: end-users may lack visibility into design choices that affect correctness, robustness, fairness, and regulatory compliance. 
We argue that existing benchmarks are insufficient to assess whether MLE agents can be safely applied in such settings. We propose desiderata for a responsibility-centered evaluation framework and conduct an exploratory study on melanoma classification, focusing on fairness across skin tones as a responsibility constraint. When evaluating two recent MLE agents, we find that agent-generated pipelines show high variance and consistently underperform manually designed baselines in both predictive quality and fairness, despite fairness-oriented prompts. These preliminary results suggest that further research is needed towards redesigning MLE agents to allow humans to guide the search process and reliably assess the compliance and quality of the generated ML pipelines.
\end{abstract}

\maketitle

\pagestyle{\vldbpagestyle}

\ifdefempty{\vldbavailabilityurl}{}{
\vspace{.3cm}
\begingroup\small\noindent\raggedright\textbf{PVLDB Artifact Availability:}\\
The source code, data, and/or other artifacts have been made available at \url{\vldbavailabilityurl}.
\endgroup
}

\section{Introduction} 
\label{sec:intro}

Machine learning (ML) is increasingly used to automate impactful decisions, and the risks arising from this widespread use
are garnering attention from policy makers, scientists, and the media~\cite{stoyanovich2022responsible}. The resulting applications are often brittle with respect to their
input data, which leads to concerns about their correctness, reliability, and fairness~\cite{schelter2020taming,grafberger2022data,guha2024automated,karlavs2025navigating}.

\header{The rise and promises of machine learning engineering agents} Designing, implementing and evaluating end-to-end ML pipelines for real-world decision making systems is tedious and requires a high level of technical expertise~\cite{stoyanovich2022responsible}. The recent progress in the code generation capabilities of large language models steered by agentic harnesses lead to a new class of agentic systems referred to as machine learning engineering (MLE) agents~\cite{aide2025,fang2025mlzero,nam2026mle,fathollahzadeh2025catdb}. These agents autonomously generate and optimize ML pipelines for a prediction task via agentic pipeline search~\cite{phani2026stratum}. They take raw data and a task description in natural language as input, and mimic the iterative process of a data scientist implementing, evaluating and improving their ML pipeline, with the search guided by the candidate pipeline's predictive performance on held-out data. The output of the MLE agent usually consists of an executable Python script, representing the highest performing pipeline found, together with this pipeline's predictions.

\header{The responsibility gap} Marketing slogans for MLE agents promise full automation, claiming to transform ``raw data into ready-to-use models and prediction outputs with minimal human intervention''~\cite{fang2025mlzero} and to ``enable individuals with limited domain expertise to address complex ML challenges effectively''~\cite{fang2025mlzero}. Making ML engineering accessible to non-technical domain experts is a worthy goal: it could democratize data work, broaden participation in the field, and allow professionals such as doctors to build ML pipelines informed by their domain knowledge~\cite{data_cascades,boulamwini_gender_shades}.

Yet decisions made during data preparation, feature engineering, model selection, and pipeline development profoundly affect not only predictive performance, but also the correctness, robustness, fairness, and interpretability of automated decision systems~\cite{stoyanovich2022responsible,shahbazi2023through,karlavs2025navigating,guha2024automated,galhotra2022dataprism,erfanian2024chameleon}. If MLE agents abstract away the development process, where and how can end-users exercise their duty of oversight over the resulting pipeline? This responsibility gap is especially concerning because people tend to overtrust machine-generated outputs~\cite{automation_bias}. In sensitive and regulated domains such as medicine, insufficient oversight is particularly problematic: ML pipelines must satisfy not only performance goals, but also standards of safety, accountability, and fairness~\cite{eu_ai_act_2024,fda2025artificial}.

\header{The need to evaluate the responsibility properties of MLE agents} The strong marketing claims for MLE agents are accompanied by impressive results on MLE-bench~\cite{chan2024mle,toledo2025ai}, a popular OpenAI benchmark derived from Kaggle competitions. MLE-bench evaluates agents in the controlled setting of predictive modeling competitions, yet it neither tests whether generated pipelines satisfy regulatory compliance requirements nor measures what level of technical expertise is required to generate such pipelines successfully. This leaves open whether MLE agents perform reliably in real-world settings, and what level of expertise domain experts need to apply them safely.
This limitation is especially important in sensitive and regulated domains such as healthcare, which account for a substantial share of MLE-bench: 14 of its 75 Kaggle competitions are medical tasks. We therefore ask: {\em Can existing MLE agents adhere to responsibility constraints when used by non-technical domain experts in sensitive application areas?}

\header{Overview and contributions} In this exploratory work, we take a first step toward addressing this research question.

\headerul{Contribution 1: Desiderata for a new evaluation framework} We outline the desiderata for a new evaluation framework for the responsibility properties of MLE agents in sensitive domains. Then we assess the widely-used MLE-bench benchmark against them (\Cref{sec:problem}).

\headerul{Contribution 2: Exploratory experiment} Based on these desiderata, we evaluate two recent MLE agents on a melanoma classification task under the responsibility constraint of classifier fairness across skin tones (\Cref{sec:approach}). Our preliminary findings are concerning: agent-generated pipelines show high variance and consistently underperform manually designed baselines in both predictive quality and fairness, despite being explicitly prompted to produce fair outputs (\Cref{sec:evaluation}).

\headerul{Contribution 3: Open artifacts} We release our code, detailed experimental logs and agentic trajectories under an open license at \textcolor{blue}{\url{https://github.com/anna-richter/be-fair}}.

\vspace{1mm}\noindent We outline our directions for follow-up research in \Cref{sec:next}.

\section{Responsibility-Centered Evaluation of MLE Agents}
\label{sec:problem}
We propose desiderata for the responsibility-centered evaluation  of MLE agents and assess MLE-bench against them.

\header{Desiderata for a new evaluation framework} Our desiderata are as follows:

\headerl{Domain-centric evaluation design} Evaluation tasks should preserve the real-world complexity of the application area. Rather than optimizing for a single metric, agents should be evaluated on their ability to account for multiple domain-specific objectives and trade-offs. This mirrors broader critiques of fairML benchmarking, which argue that intrinsic, single-metric evaluation strips away the normative context in which fairness harms materialize~\cite{pechenizkiy2025benchmarking}. Moreover, performance should be measured on an independent test set unseen by the agent, and compared against peer-reviewed expert pipelines and human expert decisions.

\headerl{Adherence to responsibility constraints} The evaluation framework should quantify the extent to which agent-generated pipelines satisfy relevant responsibility constraints, such as fairness requirements. This requires defining task-specific criteria and conducting a data-centric analysis of the generated pipelines, considering data access patterns, the variability of outputs, and the existence of fairness interventions.

\headerl{Impact of technical expertise level} Finally, the evaluation framework should assess whether MLE agents are usable by domain experts with limited technical expertise, such as medical doctors or human resources professionals. In particular, it should measure the quality agents achieve out of the box, without technically refined prompting or debugging, and determine what level of refinement is needed to obtain high-quality results. This requires varying the specificity and technical detail of the task instructions given to the agent.

\header{MLE-bench and its limitations} Most recently proposed MLE agents, including \cite{aide2025,fang2025mlzero,toledo2025ai,nam2026mle}, are evaluated on OpenAI's MLE-bench~\cite{chan2024mle}, a benchmark derived from 75 Kaggle competitions. Agents receive raw data and a natural language task description; success is measured by Kaggle's ``medal winning rate,'' i.e., how often an agent places in the top decile of competition submissions. MLE-bench spans a wide range of domains, including 14 tasks from the highly regulated medical domain~\cite{eu_ai_act_2024,fda2025artificial}. Measured against the desiderata above, however, it has several conceptual limitations.

\headerl{Limited relevance of the benchmark metric} It is unclear whether placing in the top decile of Kaggle submissions translates to sufficient quality in sensitive real-world settings.

\headerl{Lack of multi-objective optimization} Kaggle solutions are typically ranked by a single competition-specific performance score. This setup does not reveal whether agents can reliably optimize pipelines for multiple objectives, such as accuracy and fairness.

\headerl{Lack of metadata for responsibility assessment} Although MLE-bench includes 14 medical use cases, it often lacks the metadata needed to assess responsibility properties. For example, the SIIM-ISIC Melanoma Classification task~\cite{siim-isic-melanoma-classification} does not provide skin-tone annotations, making it impossible to evaluate the fairness of an agent-generated pipeline with respect to skin tone.
\section{Exploratory Experiment} 
\label{sec:approach}

\begin{figure*}[t!]
    \centering
    \includegraphics[trim={0 0 5cm 0},clip,width=\textwidth]{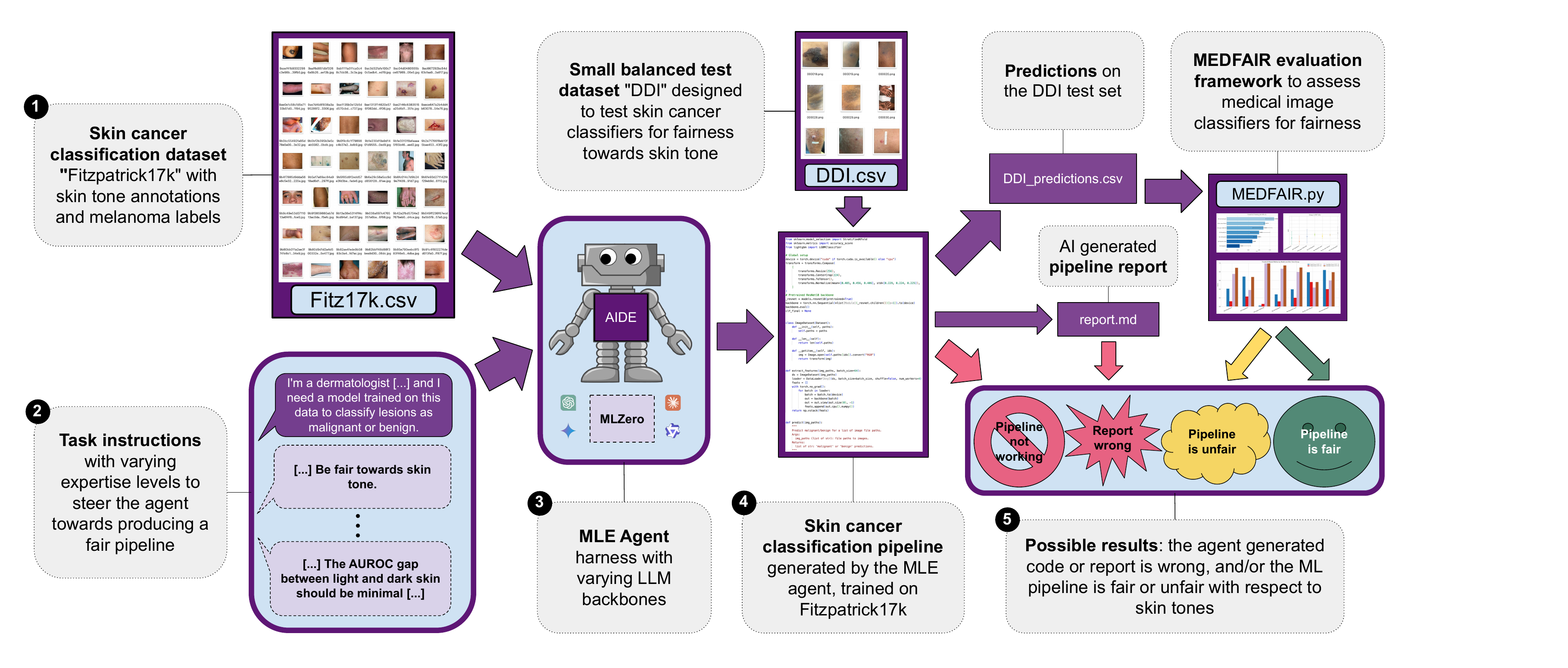}
    \caption{Overview of our exploratory experiment. \protect\circled{1} A dataset about melanoma classification for skin cancer detection (with skin tone annotations), combined with \protect\circled{2} natural language task instructions of varying technical expertise levels is given to an \protect\circled{3} MLE agent. The agent generates an ML pipeline \protect\circled{4} and an accompanying report for the task, which are subsequently evaluated \protect\circled{5} for correctness, predictive performance and fairness, and compared to manually designed expert baselines.}
    \label{fig:overview}
\end{figure*}

Based on the outlined desiderata, we design an example evaluation task in the highly regulated medical domain~\cite{eu_ai_act_2024,fda2025artificial}, focusing on melanoma classification for skin cancer detection. As a responsibility constraint, we require pipelines to provide comparable detection performance for individuals with different skin tones. \Cref{fig:overview} provides an overview of the experimental design.

\headerl{Curated datasets with rich metadata} We use ``Fitzpatrick17k''~\cite{fitzpatrick17k} as the training dataset~\circled{1}, which contains over 17,000 images with skin-tone annotations based on the Fitzpatrick scale (a categorization of skin tones by their response to ultraviolet light). To reduce the risk that agents rely on memorized solutions, we obfuscate the dataset by renaming and shuffling columns and removing external URLs. As an independent test set, we use the ``Diverse Dermatology Images (DDI)''~\cite{DDI} dataset, which is designed to evaluate skin-lesion classifiers for fairness across skin tones. DDI contains 656 images balanced across light and dark skin tones, with matched patient characteristics and diagnoses. It also includes baseline decisions from a panel of dermatologists who manually classified the images.

\headerl{Task instructions of varying expertise} We provide each MLE agent with one of four task instructions~\circled{2}, designed to reflect increasing levels of technical expertise and specification detail.
\begin{itemize}[noitemsep,leftmargin=*]
\item \texttt{base} provides a colloquial task description: \textit{`I'm a dermatologist. My colleagues and I have curated a skin lesion dataset over the past several years. I need a model trained on this data to classify lesions as malignant or benign.\footnote{For technical reasons, we add one sentence to enable later pipeline evaluation: \textit{`The trained model should be saved, and I need a function that takes a folder path of new images and returns a malignancy probability (0--1) for each image.''}}''}
\item \texttt{fairness-hint} adds the instruction \textit{`Be fair towards skin tone.''}
  \item \texttt{fairness-metric} additionally specifies the fairness objective: \textit{`The AUROC gap between light and dark skin should be minimal, without compromising overall model performance.''}
\item \texttt{fairness-methods} additionally suggests possible fairness interventions: \textit{``To improve fairness across skin tones, you may apply techniques such as filtering, upsampling, group reweighting in the loss function, or any other appropriate methods.''}
\end{itemize}

\headerl{Agentic pipeline generation} We provide the training data and one selected task instruction to an MLE agent~\circled{3}, which then executes its search and optimization procedure to generate a classification pipeline as a Python script, potentially accompanied by a 
report.

\headerl{Evaluation metrics and expert comparisons} We evaluate the agent-generated pipelines~\circled{4} on DDI, following the peer-reviewed MEDFAIR study~\cite{zong2022medfair}. We reuse its code and metrics~\circled{5}: classification performance is measured by AUC, and fairness by the AUC gap, defined as the difference in AUC between the best- and worst-performing skin-tone groups. We compare the agent-generated pipelines against decisions from the DDI human expert panel and three human-written expert pipelines. 

The expert pipelines apply ``minimax pareto selection''~\cite{martinez2020pareto} (a Pareto-efficient choice that minimizes the maximum downside across the overall AUC and the AUC gap across skin-tone groups) to decide on early stopping during training and the selection of the model checkpoint to return. They differ in the data cleaning strategies they employ: the first pipeline \texttt{MEDFAIR} follows~\cite{zong2022medfair} and removes images with missing Fitzpatrick labels, binarizes labels into malignant versus all other classes, resizes images during preprocessing, and fine-tunes a ResNet-18 model for 30 epochs. 
The second variant \texttt{MEDFAIR-dedup} trains the same pipeline on a cleaned version of Fitzpatrick17k~\cite{abhishek2025investigating} with duplicate and noisy images removed, while the third variant \texttt{MEDFAIR-filtered} further filters out images showing non-neoplastic conditions, which are not relevant to the binary classification task.

\section{Preliminary Results} 
\label{sec:evaluation}

We present preliminary results for the exploratory experiment outlined in the previous section.

\header{Experimental setup} We evaluate the MLE agent AIDE~\cite{aide2025} (the original winner in MLE-bench) and the more recent AutoGluon-based agent MLZero~\cite{fang2025mlzero} from Amazon. We give both a search budget of 20 steps, access to an A100 GPU, and a maximum execution time of one hour per step. We use the default configuration for AIDE which applies a combination of OpenAI's \texttt{gpt-4.1}, \texttt{gpt-4.1-mini}, and \texttt{o4-mini} models with a temperature of 0.5, and use OpenAI's \texttt{gpt-4.1} as LLM backbone for MLZero with the default temperature of 0.1. We run both agents seven times for each of the four task instructions (28 runs per agent in total). We report the mean and standard deviation of the prediction quality in terms of AUC and the fairness in terms of the AUC gap. 

\headerl{Baselines} As discussed in~\Cref{sec:approach}, we compare the agent-generated pipelines against several expert baselines. Most importantly, we include the expert decisions from~\cite{DDI} (referred to as \texttt{Dermatologists}), where a panel of three dermatologists manually classified the data. We also include the expert-written pipelines \texttt{MEDFAIR}, \texttt{MEDFAIR-dedup}, and \texttt{MEDFAIR-filtered}.

\begin{figure}[t!]
    \centering
    \includegraphics[width=\columnwidth]{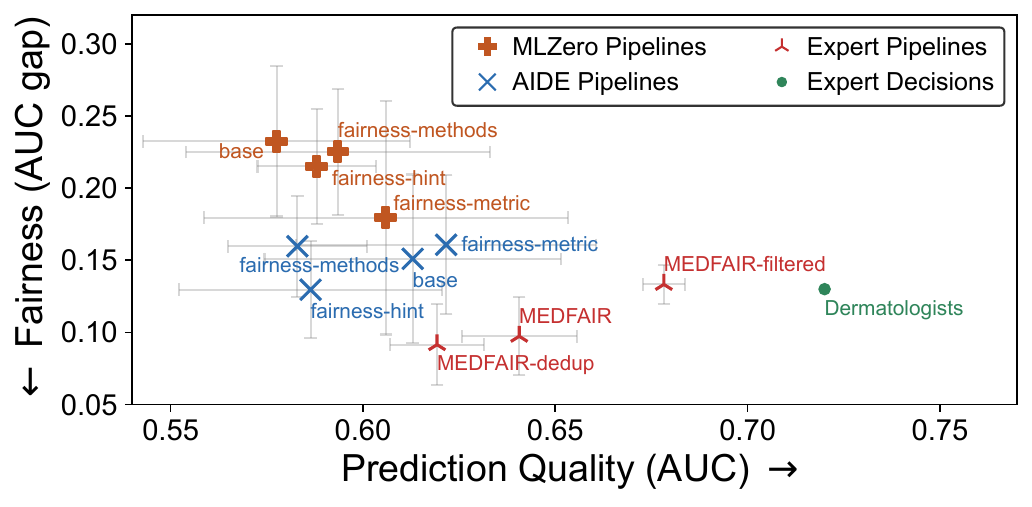}
    \caption{Prediction quality (AUC, higher is better) and fairness (AUC gap, lower is better) in the skin cancer detection task for agent-generated pipelines, manually written expert pipelines and expert decisions from dermatologists. Agent-generated pipelines are strictly dominated by expert solutions in terms of both quality and fairness, and at the same time exhibit a much higher variance in their results.}
    \label{fig:quality-vs-fairness}
\end{figure}

\header{Results and discussion} We plot the results in \Cref{fig:quality-vs-fairness}. AIDE achieves higher AUC scores and a lower fairness gap than MLZero, despite using substantially fewer tokens: approximately 6M versus 86M. However, we find that the agent-generated pipelines from AIDE and MLZero are strictly dominated by expert solutions in terms of both quality and fairness, and at the same time exhibit a much higher variance in their results. 

Furthermore, agents were unable to translate the specific instructions about fairness towards skin tone (in \texttt{fairness-metrics} and \texttt{fairness-methods}) into actual fairness gains. All AIDE and MLZero pipelines score at least 10 points lower in AUC than expert decisions, and all agent-generated pipelines are drastically outperformed in terms of fairness by the expert baseline \texttt{MEDFAIR-dedup}, which applies the outlined minimax pareto selection as fairness intervention~\cite{martinez2020pareto}.

\newpage
\headerl{Detailed analysis of the generated pipelines} We analyze the code generated by AIDE and MLZero to determine whether the pipelines monitor or improve fairness with respect to skin tone, as requested in the task instructions.
For AIDE, pipelines generated from the \texttt{base} instruction, which does not mention fairness, and the \texttt{fairness-hint} instruction, which only asks the agent to be fair toward skin tone, neither compute fairness metrics nor apply fairness interventions. They also do not access the \texttt{skin\_tone} column. Under the \texttt{fairness-metric} instruction, 3 out of 7 pipelines correctly monitor and attempt to improve fairness with respect to skin tone. Under the highly specific \texttt{fairness-methods} instruction, all pipelines include fairness monitoring and interventions, often by reweighting samples according to skin-tone frequency. However, these interventions do not translate into improved fairness, as illustrated in~\Cref{fig:quality-vs-fairness}. Manual inspection of the pipeline's intermediate results showed that the darkest group’s AUC was often close to random, suggesting that sample reweighting cannot remedy the group’s weak discriminatory performance.
For MLZero, about half of the pipelines generated from the \texttt{base} instruction pass \texttt{skin\_tone} to AutoGluon as a tabular feature, but none apply explicit fairness interventions. Pipelines from the \texttt{fairness-hint} instruction also use \texttt{skin\_tone} as a tabular feature and add class reweighting (independent of skin tone) as a fairness measure. The \texttt{fairness-metric} pipelines are similar, although two additionally report per-group AUC. Under the highly specific \texttt{fairness-methods} instruction, 2 of 7 pipelines implement explicit skin-tone-group reweighting, which again is accompanied by a close to random model performance on the the darkest group’s samples and therefore does not result in fairness gains.

\headerl{Hallucinated reports and execution bugs} When experimenting with AIDE we encountered severe bugs and had to manually patch the framework to ensure that generated pipelines were actually executed. In particular, many generated scripts failed because they relied on an \texttt{if \_\_name\_\_ == \_\_main\_\_:} block but were launched in subprocesses without the correct ``main'' context. As a consequence, some runs completed without executing any pipeline successfully, yet the agent still produced final reports containing hallucinated summaries and fabricated performance metrics.

\section{Conclusion \& Next Steps} 
\label{sec:next}

Our exploratory study suggests that agent-generated pipelines underperform expert-designed baselines in both predictive quality and fairness, while also exhibiting substantial variance and implementation flaws. In our setting, prompting agents to produce fairer pipelines was not sufficient to reliably improve fairness outcomes. 

These findings point to a  mismatch between the automation promises of current MLE agents and their readiness for high-stakes settings. This motivates future work on a more comprehensive evaluation framework based on the desiderata outlined in \Cref{sec:problem}. We plan to cover a broader range of high-stakes scenarios, complement benchmark scores with extrinsic, context-sensitive assessment~\cite{pechenizkiy2025benchmarking} and evaluate different MLE agents, backing LLMs, and search policies.  In the long term, such a framework will help guide research toward designing MLE agents that allow humans to meaningfully steer the search process and assess the compliance and quality of generated ML pipelines.

\vspace{1mm}\small{
\noindent\textit{The authors acknowledge the Scientific Computing of the IT Division at the Charité - Universitätsmedizin Berlin for providing computational resources that have contributed to the research results reported in this paper.}}



\begin{thebibliography}{19}

\bibitem[\protect\citeauthoryear{Abhishek et~al\mbox{.}}{Abhishek et~al\mbox{.}}{}]{abhishek2025investigating}
\bibfield{author}{\bibinfo{person}{Abhishek} {et~al\mbox{.}}}
\newblock Investigating the Quality of DermaMNIST and Fitzpatrick17k Dermatological Image Datasets.
\newblock \bibinfo{journal}{\emph{Scientific Data}} \bibinfo{volume}{12}, \bibinfo{number}{1}, 2025.
\newblock

\bibitem[\protect\citeauthoryear{Boulamwini et~al\mbox{.}}{Boulamwini et~al\mbox{.}}{}]{boulamwini_gender_shades}
\bibfield{author}{\bibinfo{person}{Boulamwini} {et~al\mbox{.}}}
\newblock Gender Shades: Intersectional Accuracy Disparities in Commercial Gender Classification.
\newblock \bibinfo{booktitle}{\emph{FAccT}'18}.
\newblock

\bibitem[\protect\citeauthoryear{Chan et~al\mbox{.}}{Chan et~al\mbox{.}}{}]{chan2024mle}
\bibfield{author}{\bibinfo{person}{Chan} {et~al\mbox{.}}}
\newblock {MLE}-bench: Evaluating Machine Learning Agents on Machine Learning Engineering.
\newblock \bibinfo{booktitle}{\emph{ICLR}'25}.
\newblock

\bibitem[\protect\citeauthoryear{Daneshjou et~al\mbox{.}}{Daneshjou et~al\mbox{.}}{}]{DDI}
\bibfield{author}{\bibinfo{person}{Daneshjou} {et~al\mbox{.}}}
\newblock Disparities in dermatology {AI} performance on a diverse, curated clinical image set.
\newblock \bibinfo{journal}{\emph{Science Advances}} \bibinfo{volume}{8}, \bibinfo{number}{32}, 2022.
\newblock


\bibitem[\protect\citeauthoryear{Erfanian et~al\mbox{.}}{Erfanian et~al\mbox{.}}{}]{erfanian2024chameleon}
\bibfield{author}{\bibinfo{person}{Erfanian} {et~al\mbox{.}}}
\newblock Chameleon: Foundation Models for Fairness-aware Multi-modal Data Augmentation to Enhance Coverage of Minorities.
\newblock \bibinfo{journal}{\emph{VLDB}'24}.
\newblock


\bibitem[\protect\citeauthoryear{Fang et~al\mbox{.}}{Fang et~al\mbox{.}}{}]%
        {eu_ai_act_2024}
\newblock EU AI Act, Regulation 2024/1689, https://eur-lex.europa.eu/eli/reg/2024/1689/oj.

\bibitem[\protect\citeauthoryear{Fang et~al\mbox{.}}{Fang et~al\mbox{.}}{}]{fang2025mlzero}
\bibfield{author}{\bibinfo{person}{Fang} {et~al\mbox{.}}}
\newblock Mlzero: A multi-agent system for end-to-end machine learning automation.
\newblock \bibinfo{journal}{\emph{NeurIPS}'25}.
\newblock

\bibitem[\protect\citeauthoryear{Fathollahzadeh et~al\mbox{.}}{Fathollahzadeh et~al\mbox{.}}{}]{fathollahzadeh2025catdb}
\bibfield{author}{\bibinfo{person}{Fathollahzadeh} {et~al\mbox{.}}}
\newblock CatDB: Data-Catalog-Guided, LLM-Based Generation of Data-Centric ML Pipelines.
\newblock \bibinfo{journal}{\emph{VLDB}'25}.
\newblock

\bibitem[\protect\citeauthoryear{FDA}{FDA}{}]{fda2025artificial}
\bibfield{author}{\bibinfo{person}{FDA}.}
\newblock Artificial Intelligence-Enabled Device Software Functions: Lifecycle Management and Marketing Submission Recommendations.
\newblock

\bibitem[\protect\citeauthoryear{Galhotra et~al\mbox{.}}{Galhotra et~al\mbox{.}}{}]{galhotra2022dataprism}
\bibfield{author}{\bibinfo{person}{Galhotra} {et~al\mbox{.}}}
\newblock Dataprism: Disconnect between data and systems.
\newblock \bibinfo{journal}{\emph{SIGMOD}'22}.
\newblock

\bibitem[\protect\citeauthoryear{Grafberger et~al\mbox{.}}{Grafberger et~al\mbox{.}}{}]{grafberger2022data}
\bibfield{author}{\bibinfo{person}{Grafberger} {et~al\mbox{.}}}
\newblock Data distribution debugging in ML pipelines.
\newblock \bibinfo{journal}{\emph{VLDBJ}'21}.
\newblock

\bibitem[\protect\citeauthoryear{Groh et~al\mbox{.}}{Groh et~al\mbox{.}}{}]{fitzpatrick17k}
\bibfield{author}{\bibinfo{person}{Groh} {et~al\mbox{.}}}
\newblock Evaluating deep neural networks trained on clinical images in dermatology with the fitzpatrick 17k dataset.
\newblock \bibinfo{journal}{\emph{CVPR}'21}.
\newblock



\bibitem[\protect\citeauthoryear{Guha et~al\mbox{.}}{Guha et~al\mbox{.}}{}]{guha2024automated}
\bibfield{author}{\bibinfo{person}{Guha} {et~al\mbox{.}}}
\newblock Automated data cleaning can hurt fairness in machine learning-based decision making.
\newblock \bibinfo{journal}{\emph{TKDE}'23}.
\newblock


\bibitem[\protect\citeauthoryear{Jiang et~al\mbox{.}}{Jiang et~al\mbox{.}}{}]{aide2025}
\bibfield{author}{\bibinfo{person}{Jiang} {et~al\mbox{.}}}
\newblock AIDE: AI-Driven Exploration in the Space of Code.
\newblock \bibinfo{journal}{\emph{arXiv:2502.13138}}.

\bibitem[\protect\citeauthoryear{Jovchevski et~al\mbox{.}}{Jovchevski  et~al\mbox{.}}{}]{automation_bias}
\bibfield{author}{\bibinfo{person}{Jovchevski } {et~al\mbox{.}}}
\newblock What is Wrong With Automation Bias?.
\newblock \bibinfo{journal}{\emph{Phil. \& Tech.}}'26.
\newblock

\bibitem[\protect\citeauthoryear{Karla{\v{s}} et~al\mbox{.}}{Karla{\v{s}} et~al\mbox{.}}{}]{karlavs2025navigating}
\bibfield{author}{\bibinfo{person}{Karla{\v{s}}} {et~al\mbox{.}}}
\newblock Navigating data errors in ML pipelines.
\newblock \bibinfo{booktitle}{\emph{SIGMOD}'25}.
\newblock


\bibitem[\protect\citeauthoryear{Martinez et~al\mbox{.}}{Martinez et~al\mbox{.}}{}]{martinez2020pareto}
\bibfield{author}{\bibinfo{person}{Martinez {et~al\mbox{.}}}
\newblock Minimax pareto fairness: A multi objective perspective}
\newblock \bibinfo{booktitle}{\emph{ICML}'20}.
\newblock


\bibitem[\protect\citeauthoryear{Nam et~al\mbox{.}}{Nam et~al\mbox{.}}{}]{nam2026mle}
\bibfield{author}{\bibinfo{person}{Nam} {et~al\mbox{.}}}
\newblock Mle-star: Machine learning engineering agent via search and targeted refinement.
\newblock \bibinfo{journal}{\emph{NeurIPS}'25}.
\newblock

\bibitem[\protect\citeauthoryear{Pechenizkiy et~al\mbox{.}}{Pechenizkiy et~al\mbox{.}}{}]{pechenizkiy2025benchmarking}
\bibfield{author}{\bibinfo{person}{Pechenizkiy} {et~al\mbox{.}}}
\newblock From Benchmarking to Understanding FairML.
\newblock \bibinfo{journal}{\emph{ECAI}'25}.
\newblock

\bibitem[\protect\citeauthoryear{Phani et~al\mbox{.}}{Phani et~al\mbox{.}}{}]{phani2026stratum}
\bibfield{author}{\bibinfo{person}{Phani} {et~al\mbox{.}}}
\newblock stratum: A System Infrastructure for Massive Agent-Centric ML Workloads.
\newblock \bibinfo{journal}{\emph{arXiv:2603.03589}}.
\newblock

\bibitem[\protect\citeauthoryear{Sambasivan et~al\mbox{.}}{Sambasivan et~al\mbox{.}}{}]{data_cascades}
\bibfield{author}{\bibinfo{person}{Sambasivan} {et~al\mbox{.}}}
\newblock “Everyone wants to do the model work, not the data work”:  Data Cascades in High-Stakes AI.
\newblock \bibinfo{journal}{\emph{CHI}'25}.
\newblock

\bibitem[\protect\citeauthoryear{Shahbazi et~al\mbox{.}}{Shahbazi et~al\mbox{.}}{}]{shahbazi2023through}
\bibfield{author}{\bibinfo{person}{Sambasivan} {et~al\mbox{.}}}
\newblock Through the fairness lens: Experimental analysis and evaluation of entity matching.
\newblock \bibinfo{journal}{\emph{VLDB}'23}.
\newblock


\bibitem[\protect\citeauthoryear{Schelter et~al\mbox{.}}{Schelter et~al\mbox{.}}{}]{schelter2020taming}
\bibfield{author}{\bibinfo{person}{Schelter} {et~al\mbox{.}}}
\newblock Taming Technical Bias in ML Pipelines
\newblock \bibinfo{journal}{\emph{IEEE DEBull}'20}.
\newblock

\bibitem[\protect\citeauthoryear{Stoyanovich et~al\mbox{.}}{Stoyanovich et~al\mbox{.}}{}]{stoyanovich2022responsible}
\bibfield{author}{\bibinfo{person}{Stoyanovich} {et~al\mbox{.}}}
\newblock Responsible data management.
\newblock \bibinfo{journal}{\emph{Comm. ACM}} \bibinfo{volume}{65}, \bibinfo{number}{6}.
\newblock

\bibitem[\protect\citeauthoryear{Toledo et~al\mbox{.}}{Toledo et~al\mbox{.}}{}]{toledo2025ai}
\bibfield{author}{\bibinfo{person}{Toledo} {et~al\mbox{.}}}
\newblock {AI} Research Agents for Machine Learning: Search, Exploration, and Generalization in {MLE}-bench.
\newblock \bibinfo{journal}{\emph{NeurIPS}'25}.
\newblock

\bibitem[\protect\citeauthoryear{Zawacki et~al\mbox{.}}{Zawacki  et~al\mbox{.}}{}]{siim-isic-melanoma-classification}
\bibfield{author}{\bibinfo{person}{Zawacki } {et~al\mbox{.}}}
\newblock SIIM-ISIC Melanoma Classification 2020, Kaggle.
\newblock \bibinfo{journal}{\emph{https://kaggle.com/competitions/siim-isic-melanoma-classification}}.
\newblock

\bibitem[\protect\citeauthoryear{Zong et~al\mbox{.}}{Zong et~al\mbox{.}}{}]{zong2022medfair}
\bibfield{author}{\bibinfo{person}{Zong} {et~al\mbox{.}}}
\newblock MEDFAIR: benchmarking fairness for medical imaging.
\newblock \bibinfo{journal}{\emph{ICLR}'22}.
\newblock

\end{thebibliography}



\end{document}